**Deep learning for multi-label classification of coral conditions in the Indo-Pacific via underwater photogrammetry**

Xinlei Shao[1, *], Hongruixuan Chen[1], Kirsty Magson[2], Jiaqi Wang[1], Jian Song[1], Jundong Chen[1], Jun Sasaki[1]

[1]Graduate School of Frontier Sciences, The University of Tokyo, Kashiwanoha, Kashiwa City, Chiba, 277-8563, Japan

[2]New Heaven Reef Conservation Program, Koh Tao, Surat Thani, 84360, Thailand

**Correspondence**

Xinlei Shao, Graduate School of Frontier Sciences, The University of Tokyo, Kashiwanoha, Kashiwa City, Chiba, 277-8563, Japan.

Email: 8671299838@edu.k.u-tokyo.ac.jp

**Abstract**

1. Since coral reef ecosystems face threats from human activities and climate change, coral conservation programs are implemented worldwide. Monitoring coral health provides references for guiding conservation activities. However, current labor-intensive methods result in a backlog of unsorted images, highlighting the need for automated classification. Few studies have simultaneously utilized accurate annotations along with updated algorithms and datasets.

2. This study aimed to create a dataset representing common coral conditions and associated stressors in the Indo-Pacific. Concurrently, it assessed existing classification algorithms and proposed a new multi-label method for automatically detecting coral conditions and extracting ecological information.

3. A dataset containing over 20,000 high-resolution coral images of different health conditions and stressors was constructed based on the field survey. Seven representative deep learning architectures were tested on this dataset, and their performance was quantitatively evaluated using the F1 metric and the match ratio. Based on this evaluation, a new method utilizing the ensemble learning approach was proposed.

4. The proposed method accurately classified coral conditions as healthy, compromised, dead, and rubble; it also identified corresponding stressors, including competition, disease, predation, and physical issues. This method can help develop the coral image archive, guide conservation activities, and provide references for decision-making for reef managers and conservationists.

5. The proposed ensemble learning approach outperforms others on the dataset, showing State-Of-The-Art (SOTA) performance. Future research should improve its generalizability and accuracy to support global coral conservation efforts.

# 1 Introduction

## 1.1 Background

The coral reef ecosystem, recognized as one of the most diverse and productive ecosystems on Earth, provides essential ecosystem services, including habitats for commercial fish, sources of medicine, tourism, and education (Sheppard, 2021). However, these ecosystems are increasingly threatened by human activities and climate change. The coral reef ecosystem at Koh Tao, a popular tourist destination in Thailand, has experienced significant deterioration due to human activities since 1992. Yeemin, Sutthacheep, & Pettongma (2006) reported that from 2002 to 2006, coral coverage in Koh Tao decreased by 17% due to tourism and economic development. Furthermore, coral-algal phase shifts driven by climate change have impacted the structure of the coral community (Scott et al., 2017).

To monitor and restore the local coral reef ecosystem, a coral conservation program named the New Heaven Reef Conservation Program (NHRCP) was established in 2007. This program sends divers to collect data on coral conditions and engage in conservation activities (New Heaven Reef Conservation Program, 2016). The strength of this program lies in its provision of ground truth data, the restoration of reefs, and the enhancement of community engagement in conservation activities (Grol et al., 2021; Hesley, Kaufman, & Lirman, 2023). However, there are challenges regarding the efficiency of the survey (Terayama et al., 2021) and the consistency of the data (Raoult et al., 2016). Data collection can be time-consuming; for instance, a volunteer diver takes approximately 80 minutes to complete a text-based disease survey of a 100-meter transect line, according to our NHRCP team in Koh Tao. Moreover, the backlog of unlabeled or unsorted photos due to limited resources further complicates the process. To address these issues, developing an automatic image identification model could be beneficial. Such a model would automatically identify and document coral conditions, thereby enhancing fieldwork efficiency. Additionally, the consistency of the data is significantly influenced by the experience and training of the divers, as they are responsible for identifying the taxonomy and condition of corals. The problem of data inconsistency could be mitigated by adopting a uniform, unbiased identification scheme executed by the computer model, as highlighted in the review of Arsad et al. (2023).

**1.2 Progression in the field of coral reef monitoring**

Many studies have been conducted in remote sensing and photogrammetry to monitor the condition of coral reef ecosystems. In particular, with the rapid advancements in computer vision techniques, many studies have employed classification, object detection, and semantic segmentation to analyze images of coral reef ecosystems.

In remote sensing, current research predominantly focuses on acquiring "coarse" ecological and geographical information on a large scale. Modasshir et al. (2018) introduced a Convolutional Neural Network (CNN) based object detection method using the RetinaNet architecture to identify and count the number of colonies of different coral forms. Kennedy et al. (2021) developed a tool named Reef Cover for large-scale geomorphic mapping and applied it to the Great Barrier Reef and Micronesia. González-Rivero et al. (2020) utilized the VGG-16 architecture (Simonyan and Zisserman, 2015) for segmentation to map and calculate the benthic composition using a global coral reef monitoring dataset. Similarly, Terayama et al. (2021) collected large-scale 2D images using a custom-developed camera array. They implemented image segmentation through U-Net (Ronneberger et al., 2015) to map the coverage of corals, seagrass, and sea urchins. However, these methods face limitations due to hardware constraints, leading to a loss of image detail. For example, the difference between a newly killed coral and a bleached coral lies in the presence of coral tissue (Scott, 2019), and the image resolution collected in current remote sensing studies is too low to distinguish between these two states. Additionally, marine biologists and conservationists are interested in understanding interspecies relationships, such as predation, competition, and parasitism between other species and coral. For example, the lack of detailed imagery datasets led to difficulties in identifying small, cryptic predators such as *Drupella* Snails. Therefore, in situ, high-resolution close images are needed to gain a more comprehensive understanding of the coral condition. Such detailed imagery can effectively complement the broader, large-scale maps obtained through remote sensing.

In photogrammetry, the main research gaps lie in the reliance on obsolete algorithms, the use of outdated and inconsistently labeled datasets, and the lack of comprehensive ecological information provided by images. Gómez-Ríos et al. (2019) summarized that CNN architectures like Visual

Geometry Group (VGG) (Simonyan and Zisserman, 2015), Inception v3 (Szegedy et al., 2016), ResNet (He et al., 2016), and DenseNet (Huang et al., 2017) have been popular for coral image classification. Borbon et al. (2021) applied Inception v3, ResNet-50, ResNet-152, DenseNet-121, and DenseNet-161 to classify the coral condition as healthy, dead, or bleached using images from other researchers and Google. Despite these efforts, the latest algorithm they applied is DenseNet, proposed in 2017, while transformer models have been adapted to computer vision and have shown great promise nowadays (Dosovitskiy et al., 2021). Zhang, Gruen, & Li (2022) evaluated the performance of five SOTA neural network architectures to determine the spatial distribution and the living/ dead state of *Pocillopora sp.* on Moorea Island. However, their focus on a single genus and two health states (living or dead) limits the understanding of overall habitat health. Williams et al. (2019) employed CoralNet, based on the EfficientNet-B0 backbone proposed by Tan and Le (2019) and Chen et al. (2021), to estimate benthic coverage rates in the main Hawaiian Islands and American Samoa using images from 2010-2016. However, the dataset was last updated in 2016. Similarly, González-Rivero et al. (2020) applied VGG-16 and Support Vector Machine (SVM) to estimate the benthic abundance based on coral images from five different global regions in the period from 2012 to 2016. However, this temporal gap since the last update in 2016 means that the ecological information provided is potentially outdated, which could lead to inaccurate or misleading conclusions, especially since the coral reef ecosystem is very dynamic and has changed rapidly due to climate change in recent years. Apart from the obsolete datasets, the inconsistency and inaccuracy in datasets can also lead to biased results, as the label of images posted on Google can be inaccurate, and pictures collected from different sites and by various devices can lead to variation. Additionally, our NHRCP team and Gómez-Ríos et al. (2019) have highlighted that existing coral datasets seldom contain close-up photos (to reflect coral texture) and distant photos (to provide colony structure information), which are crucial for precise annotation. More importantly, some of the latest advances in computer vision, i.e., the vision Transformer architecture, their potential in this task has not yet been explored.

To bridge the gaps in existing studies and to tackle the challenges of coral condition classification, this study makes the following contributions: (1) Dataset: It developed a uniform, updated, open-source dataset with more than 20,000 high-resolution samples collected through field surveys, with close-up and distant photos of the same coral colony. (2) Algorithm: It applied, compared, and analyzed the performance of the mainstream neural networks applied in coral image classification and proposed a new multi-label classification method using the most promising SOTA models in computer vision. The ensemble learning method was applied to deal with the complex nature of ecological tasks, hence achieving a higher accuracy. (3) Ecological information: Through close collaboration with marine biologists and conservationists, this research aligned with the real-world needs of these experts. This partnership ensured the quality of the label and the professional ecological information retrieved from the classification output.

In conclusion, this research, based on fieldwork conducted on the island of Koh Tao, aims to apply SOTA computer vision algorithms to automate the identification of coral health conditions. This approach is intended to significantly enhance the efficiency and accuracy of coral monitoring and to propose a new model that contributes to more effective coral conservation efforts.

## 2 Methods

### 2.1 Study areas and fieldwork design

Koh Tao, a 21 km$^2$ island located at 10°5′24′′N and 99°50′17′′E, is on the western shore of the Gulf of Thailand. Renowned as a popular diving destination, the island is surrounded by biodiverse, dense fringing reefs (Stuij et al., 2022). This study area was selected due to its high dependence on tourism (Magson et al., 2022) and the availability of cooperation from the local NHRCP team.

The field survey covered major diving spots around Koh Tao. They were selected for favorable weather conditions during the fieldwork, such as current and visibility, and for representing typical coral habitats facing diverse challenges. At each dive site, underwater images were collected using an Olympus TG-6 camera operated by divers. Photos were taken following the convention of the local NHRCP team, where a healthy colony only contains one image. In contrast, a compromised colony includes photos of two scales: one at the colony scale and the other at a close-up scale within a 40cm range of the subject (reflected in Figure 1). The combination of different photo scales ensures an accurate diagnosis of coral health. All images were captured using natural light, with manual adjustments to the white balance as necessary to accommodate changes in diving depth.

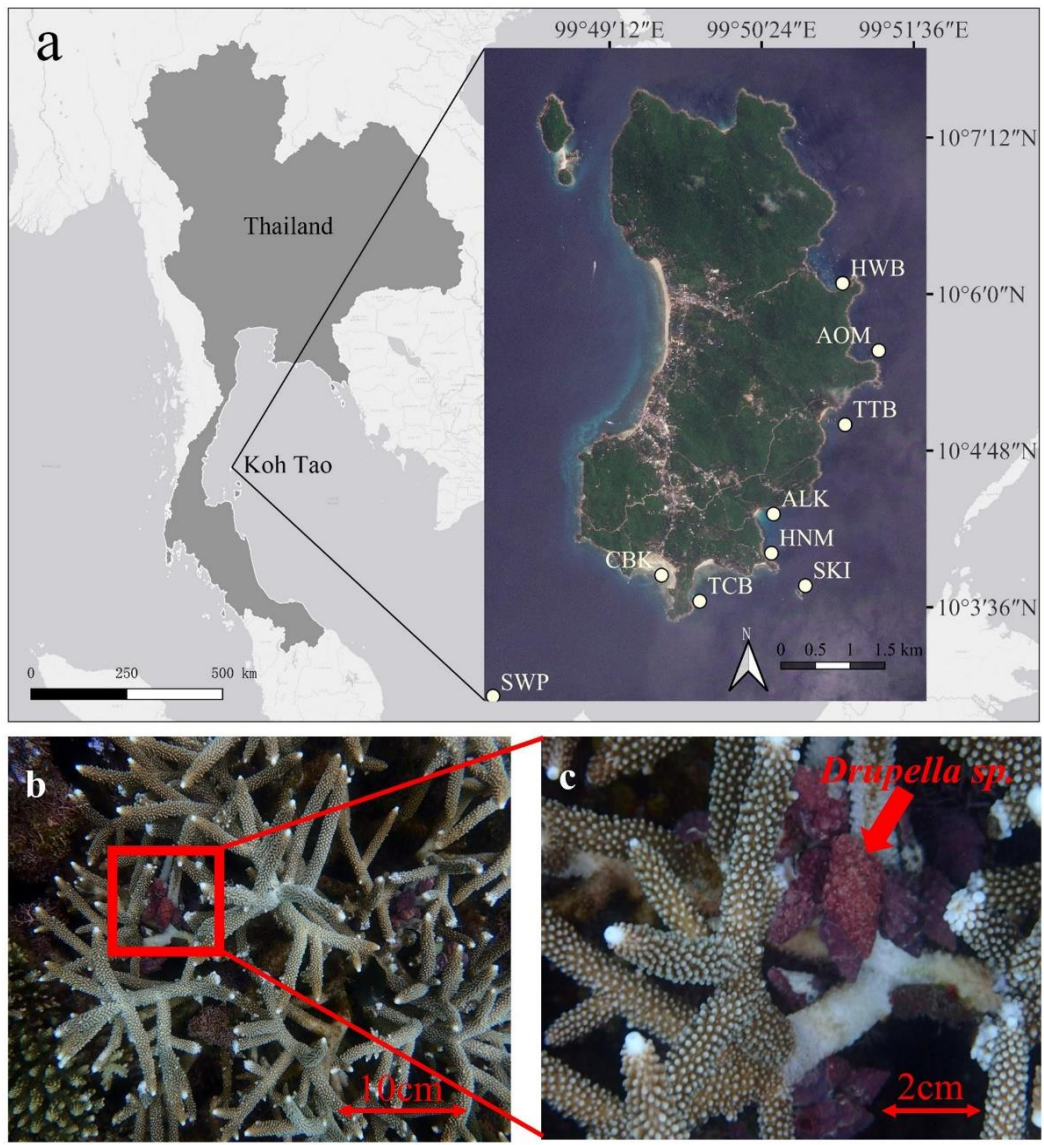

**Fig. 1.** a) Overview of the study region and diving sites covered in the field survey, where HWB = Hin Wong Bay, AOM = Aow Mao, TTB = Tanote Bay, ALK = Aow Leuk, HNM = Hin Ngam, SKI = Shark Island, TCB = Taa Chaa Bay, CBK = Chalok Baan Kao Bay, SWP = Southwest Pinnacle. b) Example of underwater image containing branching coral colonies. c) Example of the close image taken within 40cm, showing the same branching coral colony with a cluster of Drupella snails as predators (marked by the arrow) on it.

**2.2 Coral image dataset**

Field observations were conducted during two survey periods: March 2023 for preliminary research and August-September 2023 for formal research. Nine surveys were conducted across nine diving sites (shown in Figure 1). The preliminary and formal research yielded 49 and 680 images of 4000×3000 pixels, respectively. These images were cropped into 20,800 patches of 512×512 pixels and then labeled and cross-checked by experts into eight classes: healthy coral, compromised coral, dead coral, rubble, competition, disease, predation, and physical issues. The definition of each class and the labeling rules were adapted from Conservation Diver (Scott, 2019) and marine ecological knowledge (Haskin, 2022). Some representative images from each class are illustrated in Figure 2. Images from the dataset were randomly partitioned into two subsets: 0.7 for training and 0.3 for testing. In deep learning, the training dataset trains the model, allowing it to learn and adjust its parameters. The test dataset is used to test the model's performance on unseen data so that an unbiased evaluation of the final model fit can be made (Murphy, 2012).

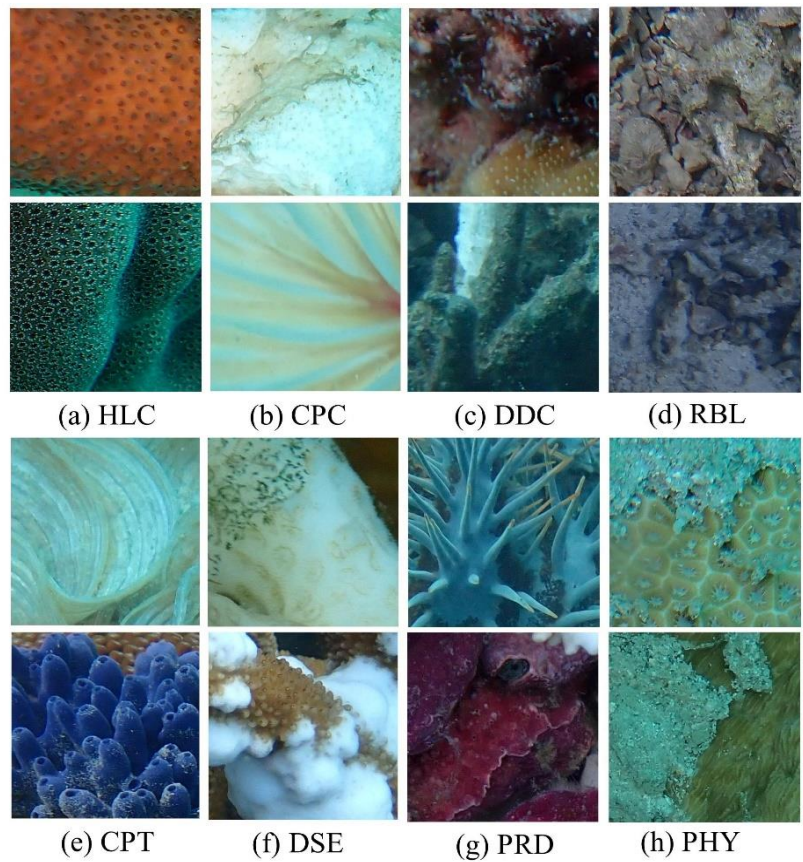

**Fig. 2.** Examples of representative patches from the dataset, where each column shows two examples per class. a) HLC = Healthy coral, b) CPC = Compromised coral, c) DDC = Dead coral, d) RBL =

Rubble, e) CPT = Competition, f) DSE = Disease, g) PRD = Predation, and h) PHY = Physical issues. Since it is a multi-label classification task, some examples contain more than one class.

## 2.3 Challenges of coral condition classification

The classification of coral conditions based on underwater photogrammetry can be challenging as it shares characteristics inherent to underwater images and ecological field photos: (1) Light variations: Light intensity varies due to the light attenuation in the water column, weather conditions, site topography, and turbidity (Edge, 2021). (2) Blurred images: Due to the complicated underwater conditions, the underwater images can be out of focus due to waves, currents, or even the imperfect buoyancy of the divers (Agrafiotis et al., 2018). (3) Irrelevant features (Lauwereyns et al., 2001): In addition to coral colonies, irrelevant features also appeared in the images, such as sea urchins, sessile animals, and fish. (4) Common traits shared across classes (Gómez-Ríos et al., 2019): Similar features are shared across classes, leading to confusion. For example, the class "rubble" shares the same textures as the class "dead coral". (5) Significant inner-class variations (Lu and Weng, 2007): The features can differ within the same class. For instance, the phenotypic expressions of a given pathogenic condition may differ significantly between coral species, resulting in different signs of the same disease across different species.

## 2.4 Deep network architecture

For coral reef monitoring, multiple instances of different coral conditions often co-occur within a single image. For instance, a coral colony can exhibit various health problems simultaneously. Therefore, this paper formulated coral condition monitoring as a multi-label classification task. Eleven mainstream neural networks, including convolutional neural networks and recently emerging Transformer architectures, were selected to tackle the challenges outlined in subsection 2.3. Through a comparative analysis of their performance, a novel multi-label classification method based on an ensemble learning approach was proposed.

*2.4.1 Visual Geometry Group (VGG)*

The VGG-16 network was introduced in 2014 by Simonyan and Zisserman (2015) and was one of the top performers in the ILSVRC 2014 competition. The model is known for its 16 weighted layers, predominantly using 3x3 convolutional filters, which allows it to better capture complex spatial hierarchies in images.

*2.4.2 ResNet*

The ResNet network, proposed by He et al. (2016), won the ILSVRC in 2015. It revolutionized deep learning by introducing residual learning with shortcut connections. This innovation enabled the training of much deeper networks without the vanishing gradient problem, while also reducing computational complexity.

*2.4.3 DenseNet*

The DenseNet network was proposed by Huang et al. (2017) and is characterized by its dense connectivity as it connects each layer to every other layer in a feed-forward fashion. The feature maps from all previous layers serve as inputs for each layer, and in turn, its feature maps are inputs for all subsequent layers. It optimizes the flow of information and gradients throughout the network, encourages feature reuse, and improves parameter efficiency.

*2.4.4 Inception v3*

The original Inception model, also known as GoogLeNet, won the ILSVRC in 2014. Inception v3 is an evolution of the original GoogLeNet (Szegedy et al., 2016). It is characterized by its efficient use of computational resources through factorized and asymmetric convolutions (such as 1x7 followed by 7x1), and it enhances model performance and training stability with expanded filter banks and batch normalization, making it a versatile architecture for complex image recognition tasks.

*2.4.5 EfficientNet*

EfficientNet is a family of convolutional neural network architectures proposed by Tan and Le (2019). It is distinguished by its compound scaling method, which uniformly scales the depth, width, and resolution based on a set of fixed coefficients. EfficientNet offers a series of scaled models (B0 to B7) that balance computational cost and performance.

*2.4.6 Vision Transformer (ViT)*

The Vision Transformer (ViT) is an approach proposed by Dosovitskiy et al. (2021), which applies transformer architecture, primarily known for its success in natural language processing (NLP), to the field of computer vision. This method represents a significant shift from traditional image processing techniques, as it does not rely on the central convolutional operations of CNNs. Instead, ViT uses transformer principles, such as attention mechanisms and positional encodings, to process and understand images.

*2.4.7 Swin Transformer*

The Swin Transformer was proposed by Liu et al. (2021), which adapts the standard transformer architecture by introducing a shifted window scheme for self-attention, partitioning images into non-overlapping windows to efficiently capture local and global features. It features a hierarchical structure that processes images on multiple scales, similar to CNNs, which enhances its applicability to a wide range of vision tasks. This architecture significantly improves computational efficiency over the original Vision Transformer, making it more scalable and effective for diverse computer vision applications.

**2.5 The ensemble learning-based approach proposed in this study**

Ensemble learning is a machine learning approach in which multiple classifiers are trained and combined to achieve better performance than any single model alone. It has shown great promise in improving the performance of image classification tasks (Chen et al., 2019). This concept has evolved and was based on methods proposed in several foundational papers, such as bagging (Breiman, 1996), boosting (Schapire, 1990), and stacking (Wolpert, 1992). In this study, two ensemble models were established: one integrated Swin-Transformer-Small, Swin-Transformer-Base, and EfficientNet-B6, while the other combined Swin-Transformer-Small, Swin-Transformer-Base, and EfficientNet-B7. It is found that these ensemble models achieved relatively higher accuracy in our classification task than the models alone. Specifically, as indicated by Figure 3b, after training the models individually, the outputs of the different models were directly summed and averaged in the inference phase to obtain more robust predictions.

**2.6 Training and optimization**

The training process for this multi-label classification task is reflected in Figure 3a. This process involves inputting images into the deep learning model and applying the binary cross-entropy loss function to each output channel. Consequently, the model can simultaneously learn and predict the presence of multiple independent labels per image.

These deep learning architectures were implemented by Pytorch. Following the standard protocol in multi-label classification tasks (Tarekegn, Giacobini, & Michalak, 2021), the binary cross-entropy loss is applied to each unit in the network output as a loss function. All models are initialized with pre-trained weights on ImageNet. All models are optimized using the AdamW optimizer (Loshchilov and Hutter, 2019) with a learning rate of $10^{-4}$, weight decay of $5*10^{-4}$, batch size of 8, and the number of training iterations is 25000.

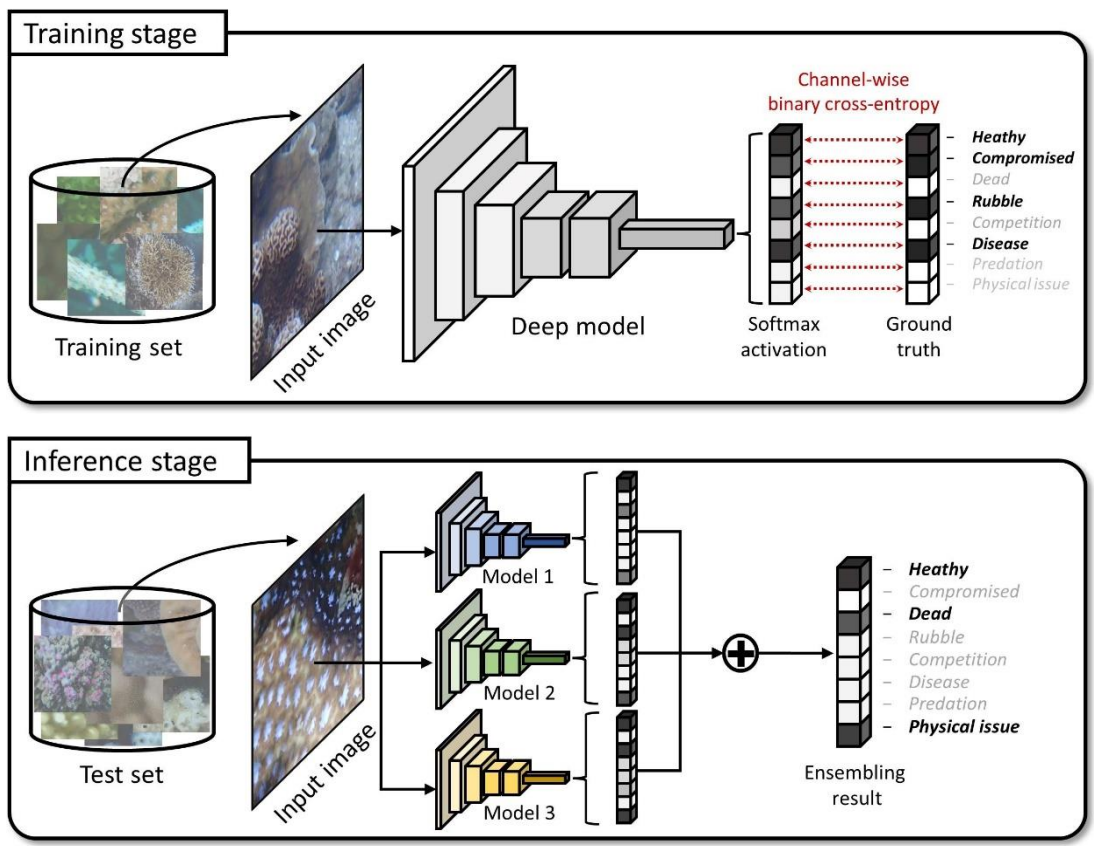

**Fig. 3.** Framework illustrating the process of developing the deep learning model for the multi-label classification task. a) Training stage: Ground truth images were imported to the deep learning model, followed by the independent application of the binary cross-entropy loss function to each output channel. b) Inference stage: test images were input to three models simultaneously, and probabilities were averaged to determine the final prediction.

## 2.7 Evaluation metrics

Classification performance was evaluated by the F1 metric, frequently used in classification tasks, especially in multi-label settings. There are two ways of calculating the F1 Score in multi-label settings: the Macro F1 Score and the Micro F1 Score (Dalianis, 2018).

The Macro F1 Score calculates the F1 score independently for each class $i$ and computes the arithmetic mean of these scores. It consists of two steps:

1. Calculate Precision, Recall, and F1 score for each class $i$, where TP = True Positive, FP = False Positive, FN = False Negative, $N$ = the total number of classes, and $i$ denotes the index of each class in the dataset:

$$Precision_i = \frac{TP_i}{TP_i + FP_i} \tag{1}$$

$$Recall_i = \frac{TP_i}{TP_i + FN_i} \tag{2}$$

$$F1_i = 2 \times \frac{Precision_i \times Recall_i}{Precision_i + Recall_i} \tag{3}$$

2. Calculate the Macro F1 Score:

$$Macro\ F1 = \frac{1}{N} \sum_{i=1}^{N} F1_i \tag{4}$$

The Micro F1 Score aggregates the total true positives, false negatives, and false positives across all classes to calculate the overall F1 Score, hence giving more weight to classes with more instances, making it more sensitive to the performance of frequent classes.

The calculation consists of two steps:

1. Calculate Precision and Recall on a global level, i.e., across all classes:

$$Precision = \frac{\sum_{i=1}^{N} TP_i}{\sum_{i=1}^{N} TP_i + \sum_{i=1}^{N} FP_i} \tag{5}$$

$$Recall = \frac{\sum_{i=1}^{N} TP_i}{\sum_{i=1}^{N} TP_i + \sum_{i=1}^{N} FN_i} \tag{6}$$

2. Calculate the Micro F1 Score:

$$Micro\ F1 = 2 \times \frac{Precision \times Recall}{Precision + Recall} \tag{7}$$

The Match Ratio is another critical parameter to assess the accuracy of the classification model. This parameter stands for the labels predicted by the model, which is precisely the same as the label of test data:

$$\text{Match Ratio} = \frac{\text{Number of Correct Predictions}}{\text{Number of Predictions}} \qquad (8)$$

**3 Results**

**3.1 Performance comparison of different tested models**

The results of 20 neural network models are presented in Table 1. These include VGG with 13, 16, and 19 layers, all incorporating batch normalization to speed up training and improve performance; ResNet with 18, 34, 50, 101, and 152 layers; DenseNet with 121, 161, and 201 layers; Inception v3; EfficientNet-B4, B5, B6, and B7; ViT-16; Swin-Transformer in tiny, small, and base variants; the ensemble model integrating Swin-S, Swin-B, and EfficientNet-B6; and the ensemble model consisting of Swin-S, Swin-B, and EfficientNet-B7.

According to Table 1, it can be found that overall, the Swin-Transformer-Base achieves the best performance compared to the other 19 models. It obtained 58.49%, 84.00%, and 78.87% of match ratio, micro F1 score, and macro F1 score, respectively. For each class, it also outperforms other models in most cases, with the only two exceptions being class "compromised coral" (CPC) and class "competition" (CPT), where Swin-Transformer-Base has a 0.36% and 4.87% lower F1 score than Swin-Transformer-Small.

When looking at the F1 score of each class separately across different models, a consistent trend emerges in the class-specific F1 scores, with certain classes consistently outperforming others regardless of the models' absolute performance metrics. For example, class "healthy coral" (HLC) has the highest F1 score among all classes. In contrast, class "rubble" (RBL), "competition" (CPT), and "physical issues" (PHY) have relatively low F1 scores compared to other classes. Specifically, for the Swin-Transformer-Base model, class "healthy coral" (HLC) (94.49%) is about 24% higher than the class "competition" (CPT) (70.52%), suggesting that the model performs worse with this class. It might be due to class imbalance, lack of representative data, confusion with other classes, and other model-tuning issues. Efforts should be put into this class to improve its performance.

**3.2 Superior performance of the ensemble models**

As illustrated in Table 1, the ensemble model, which integrates Swin-Transformer-Small, Swin-Transformer-Base, and EfficientNet-B7, improved accuracy in all classes compared to using any individual model alone. The average increase in performance is about 2% for each criterion, particularly notable with a 4.84% increase in match ratio, a significant improvement considering its previous highest value of 58.49% achieved by Swin-Transformer-Base. This result demonstrates the effectiveness of the ensemble learning approach in enhancing the model's accuracy.

**4 Discussion**

**4.1 Interpretation of the accuracy of different tested models**

According to the experiment results presented in subsection 3.2, the Swin-Transformer-Base model achieved the highest accuracy among individual models tested in this study. The key attribute accounting for the superior performance of the transformer model compared to the CNNs lies in its attention mechanism. In transformer models, the image is divided into patches, and the attention mechanism allows the model to process each patch independently and weigh each patch differently. This ability to focus attention at a finer level makes the model good at ecological tasks where fine features are critical (Kyathanahally et al., 2022). Simultaneously, the self-attention mechanism enables the model to understand how each patch relates to each other and how they relate to the overall context of the whole image. The transformer model can identify and classify target instances more accurately by detecting subtle signs of stressors via local features in each patch and capturing broader global features across the entire image. Despite little or no studies that applied transformer to coral condition monitoring, the superior performance of this model has been demonstrated in similar ecological studies such as plant disease identification (Thakur et al., 2023); insect pest classification (Peng and Wang, 2022), and underwater fish detection (Liu et al., 2024).

**4.2 Interpretation of the efficiency of different tested models**

Model efficiency is another crucial criterion for assessing a deep learning model, in addition to accuracy. A good model should achieve an optimal trade-off between these two criteria (Deng et al., 2020; Götz et al., 2022). Model efficiency is typically gauged by its number of trainable parameters, which refers to the weights and biases adjusted during the training process. Generally, higher trainable parameters indicate increased complexity, memory usage, and computational requirements (He et al., 2023).

To assess the balance between the accuracy and efficiency of the different models tested in this study, the set of trainable parameters is listed in Table 1. Within the same architecture, it is observed that the more complex the architecture, the higher the number of its parameters. For instance, as the number of convolutional layers in ResNet models increases (from 18 to 34, 50, 101, and 152), their corresponding parameters also increase (11.7M, 21.8M, 25.6M, 44.5M, and 60.2M, respectively). This increase in parameters resulted in ascending accuracy within this architecture group. This trend arises from the model's greater capacity to capture and learn intricate patterns through more parameters. However, this tendency is specific to models within the same architecture. For example, EfficientNet models (average Micro F1 score = 82.31%) outperform VGG models (average Micro F1 score = 76.53%) despite having fewer parameters (average 39.75M compared to VGG's average of 138.40M. This is attributed to their efficient architecture. Specifically, the large number of parameters in VGG models is due to the use of three fully connected layers at the end of the architecture (Basha et al., 2020). Other models, like Inception v3, have replaced these fully connected layers with Global Average Pooling (GAP) layers, significantly reducing the number of parameters while maintaining performance (Lin, Chen, & Yan, 2014, 2014). Similarly, Swin-Transformer-Base has fewer parameters than VGG models but exhibits the highest performance among all tested models. Therefore, selecting a model with high accuracy and a reasonable number of parameters is crucial.

**4.3 Analyzing the misclassified images**

This subsection will analyze the misclassified images generated by the ensemble model. By analyzing the frequency and patterns of misclassifications, classes most prone to errors and classes the model struggles to distinguish effectively can be identified. Additionally, the reasons behind these errors will be discussed. Based on these insights, corresponding optimizations can be proposed to improve the model's performance further.

As shown in Table 2, the model exhibits more False Negatives (FNs) than False Positives (FPs) across most classes, except for the healthy coral class. This tendency indicates that the model underpredicts the other seven classes while overpredicting healthy coral. This issue could be due to various factors, such as underrepresentation in the training data, lack of distinct features, or the model's threshold settings being too stringent. The balance between FNs and FPs is another important criterion for assessing the performance of the model, as it suggests whether the model has reasonably even sensitivity to detect the target instances. The dead coral class balances FN and FP best, while the competition and predation classes show significant discrepancies with higher FN. This imbalance suggests a need for an improvement in the model sensitivity to reduce missed detection rates.

In the analysis of misclassified images, it was observed that classes are predominantly misclassified as healthy coral. This pattern suggests that features indicative of compromised status or stressors may be too subtle for the model to detect accurately. As illustrated in Figure 4a, the two images on the left contain cyanobacteria, while the two on the right contain macro-algae. Even for experts, both objects are challenging to detect due to the low contrast between purple and green, and brown and red. However, human experts identified and labeled these images by referring to the unclipped large images. Therefore, it is challenging for the model to classify this type of image correctly based solely on the features of a single patch. Sedimentation belongs to the class "physical issue", Figure 4b shows coral with sediment problems but was misclassified as healthy coral. The color of the sediments in the two left images is similar to their surrounding features (the sandy bottom and *Porites sp.*), making it difficult for the model to identify. Similarities between dead coral and dead organic matter (belonging to the class physical issue) can also lead to false negative misclassification problems. In the two right images, both dead coral and dead organic matter situated on the coral are present; the model successfully identified the dead coral but failed to detect the presence of dead organic matter.

Common traits, particularly textures, shared between the class "dead coral" and "rubble" are one of the factors lowering the classification accuracy of the "rubble" class. Examples of images with rubble misclassified as dead coral are shown in Figure 4c and Figure 4d, where coral skeletons covered by filamentous algae characterize both classes. According to the classification standard adopted in this research, the significant differences between dead coral and rubble lie in whether the coral skeleton still maintains its structure. In other words, dead coral is defined as coral skeletons covered by filamentous algae but still retaining their growth forms, such as branching, encrusting, and massive (Zawada, Dornelas, & Madin, 2019). For instance, the corals shown in Figure 4d, although dead, still exhibit the same growth forms as their adjacent living parts of the colony. Their growth forms, from left to right, are branching, encrusting, encrusting, and sub-massive, respectively. In contrast, rubble refers to coral fragments that have lost their structure. For example, a coral fragment covered by algae and fallen on the encrusting coral, as shown in the left image of Figure 4c, is considered "rubble" rather than "dead" because its growth form cannot be identified from the image. In fact, without capturing the features of the structures (global information), even domain experts struggle to identify these images solely based on textures (local information). The importance of capturing global contextual information in this multi-label classification task is one of the possible reasons for the superior performance of transformer models compared to traditional CNNs.

The inherent complexity of ecological images often leads to variations within each class. According to standards applied in the coral conservation (Scott, 2019), four types of marine organisms are considered competitors of coral reefs: cyanobacteria, macro-algae, tunicate, and sponge, with significant morphological variations within the same group. Examples that show inner variations within the competition class are illustrated in Figure 4, where Figure 4e displays four types of macro-algae with completely different textures and colors, and Figure 4f shows different sponges (left two) and cyanobacteria (right two). These high variations within the class can make it challenging for the model to capture representative features, leading to misclassification. A potential solution to increase the model's performance could be a further class breakdown. However, this can be time-consuming, and typically, the first step in an ecological monitoring survey is to classify the image at a "coarse level," i.e., determining if a competitor is present in the image. Therefore, a good balance between detail and efficiency should be considered in future research.

The primary reasons for misclassification include subtle traits, common local features shared across classes, limited information from local contexts, and variations within each class. Some potential solutions to address these issues are: (1) using a larger training dataset with representative images to better enable the model to learn distinct features; (2) utilizing large-scale images instead of cutting them into 512x512 pixel patches, allowing the model to capture global context and long-distance dependencies; (3) improving the quality of underwater photos; and (4) subdividing classes with high internal variations to ensure each class has similar and uniform features. With these modifications, the classification performance is expected to be further improved.

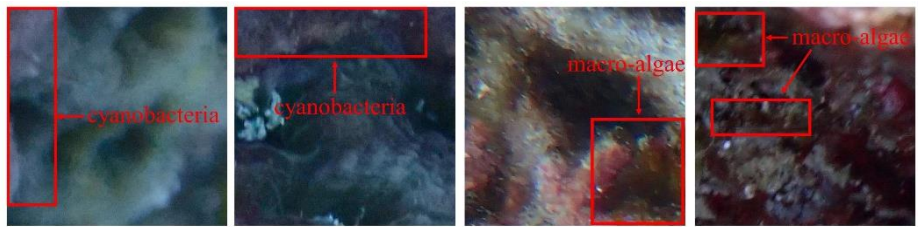

(a)

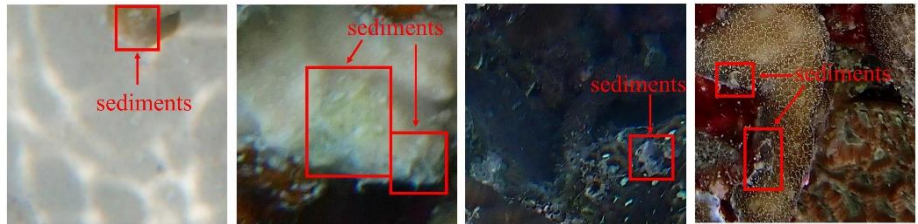

(b)

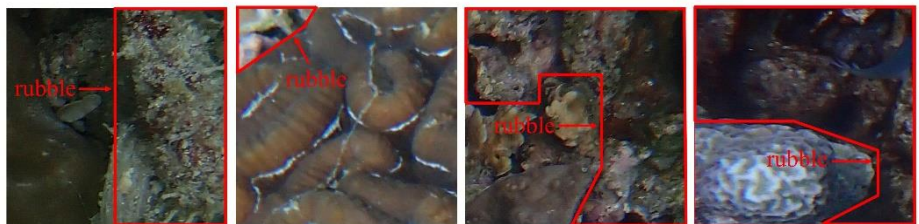

(c)

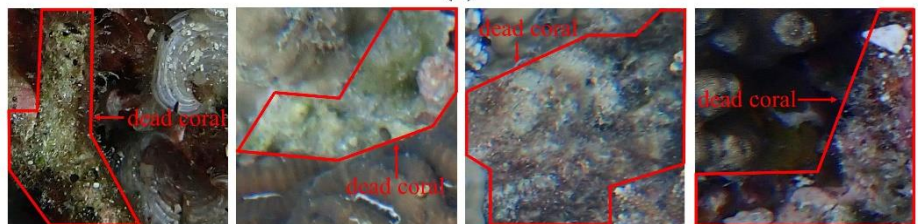

(d)

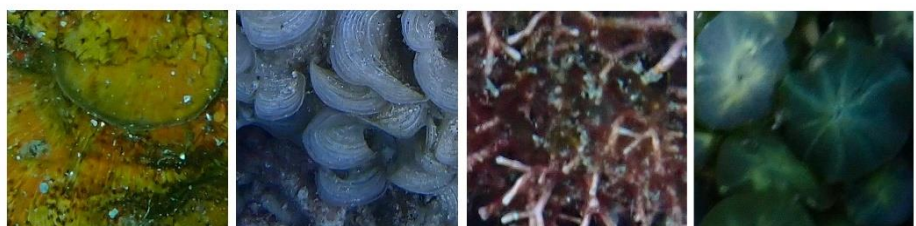

(e)

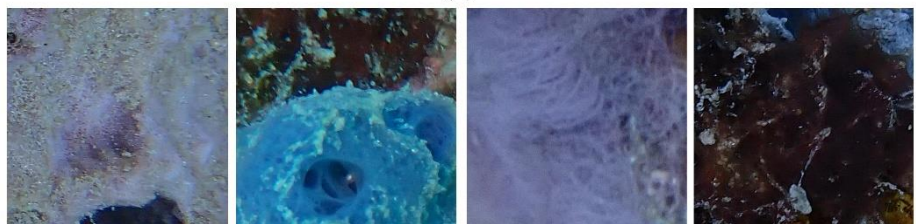

(f)

**Fig. 4.** Examples show subtle, hard-to-be-identified features that indicate the presence of (a) competition and (b) physical issues; similarities between (c) rubble and (d) dead coral; inner variations within the competition class, where (e) images of four different types of macro-algae with various morphological traits and (f) images of different sponges (left two) and cyanobacteria (right two).

**4.4 Implications for coral reef conservation**

This study proposes an ensemble model for automatic identification of the health status of corals based on underwater photogrammetry. According to the experimental results, this automated tool enables the qualitative analysis of coral conditions (healthy, compromised, dead, or rubble) and the corresponding stressors (competition, disease, predation, or physical issues) from underwater images. Coral conservation activities often involve taking large quantities of photos, such as conducting photo transect surveys of specific habitats or archiving photos, followed by manual identification, or labeling of these images. Additionally, as the local NHRCP team reported, many photos remain unlabeled due to limited human resources and time. Therefore, the application of this tool allows quick retrieval of images with specific labels by marine biologists, conservationists, and coral managers. Furthermore, since the images were collected in various habitats, the main environmental stressors that affect coral reefs in each habitat can be identified. For example, corals in areas close to human settlements, such as Chalok Baan Kao Bay, exhibit poorer health conditions and are significantly affected by sediment issues, likely due to frequent human disturbances such as nearshore construction, snorkeling, and heavy water traffic. Similarly, shallow habitats are more prone to bleaching. Consequently, valuable ecological information can be extracted from the labeled images, which helps conservation activities, such as establishing sediment traps in the corresponding habitats.

The proposed ensemble model shows great promise in its generalizations, as it can classify coral images into categories following the standard universal criteria used in coral conservation. Since the model was specifically trained using coral images from the Indo-Pacific, future research will focus on incorporating imagery from other regions. This effort aims to enhance the generalizability of the model and to test its applicability to a broader range of habitats worldwide. This model represents a starting point for quickly identifying and detecting changes in coral conditions, offering benefits to coral conservation programs and biological studies worldwide.

## 5 Conclusions

This study integrates a newly built coral image dataset with deep learning to qualitatively analyze the health condition of coral reefs. Combining Swin-Transformer-Small, Swin-Transformer-Base, and EfficientNet-B7, an ensemble model was proposed to automatically classify coral images with multiple labels, following the universal classification standard used in coral conservation programs worldwide. Consequently, our model holds great promise in enabling marine biologists and coral conservationists to identify coral conditions efficiently and objectively. The main contributions of this study are summarized as follows: (1) Generation of a high-resolution underwater image dataset comprising corals in various health conditions. (2) Comparison and evaluation of some of the most classic, SOTA deep learning models in the relevant field, revealing that the Swin-Transformer outperforms seven other tested deep learning architectures. This apparent performance advantage over SOTA convolutional neural networks hints at the importance of capturing non-local contextual information for the coral condition classification task. (3) Proposal and development of an improved multi-label classification model based on ensemble learning. This new model outperformed the other models tested in this study, achieving SOTA accuracy in this multi-label classification task. This study addresses existing research gaps by updating the dataset, applying the latest computer vision algorithms, providing professional ecological insights, and addressing practical challenges in conservation efforts. Future work will focus on collecting multi-temporal and multi-regional data to further enhance our model's generalization capability.

**ACKNOWLEDGEMENTS**

This work was supported by the Sasakawa Scientific Research Grant from The Japan Science Society (Grant/Award Number: 2023-2010). Also, this research was partially supported by the Graduate School of Frontier Sciences, The University of Tokyo, through the Challenging New Area Doctoral Research Grant (Grant/Award Number: C2301). The authors thank the New Heaven Reef Conservation Program, Koh Tao, Thailand, for assisting with the field survey, data annotation, and marine biology and conservation diving training.

## AUTHORS' CONTRIBUTIONS

**Xinlei Shao:** conceptualization (lead); data curation (lead); formal analysis (equal); funding acquisition (lead); investigation (lead); methodology (equal); project administration (lead); resources (supporting); software (supporting); writing – original draft (lead); writing – review and editing (equal). **Hongruixuan Chen:** conceptualization (supporting); data curation (equal); formal analysis (equal); methodology (lead); software (lead); writing – original draft (equal); writing – review and editing (equal). **Kirsty Magson:** data curation (equal); investigation (supporting); resources (lead); writing – original draft (supporting); writing – review and editing (equal). **Jiaqi Wang:** investigation (supporting); writing - review and editing (supporting). **Jian Song:** methodology (supporting); software (supporting); writing - review and editing (supporting). **Jundong Chen:** methodology (supporting); writing - review and editing (supporting). **Jun Sasaki:** supervision (lead); writing - review and editing (lead).

## CONFLICT OF INTEREST

All authors declare no conflict of interest.

## DATA AVAILABILITY STATEMENT

The coral image dataset and tabular data are available at https://github.com/XL-SHAO/CoralConditionDataset.

## ORCID

Xinlei Shao: 0009-0008-8652-4644

Hongruixuan Chen: 0000-0003-0100-4786

1 **Tables**

2 **TABLE 1** Performance comparison of different networks and methods. The best performance of the single and ensemble models is highlighted in bold,
3 and the second highest is stressed with underlining. The abbreviations of each class are as follows: HLC = Healthy coral, CPC = Compromised coral,
4 DDC = Dead coral, RBL = Rubble, CPT = Competition, DSE = Disease, PRD = Predation, and PHY = Physical issues.

| Model | Match Ratio | Micro F1 | Macro F1 | HLC | CPC | DDC | RBL | CPT | DSE | PRD | PHY | Parameter number |
|---|---|---|---|---|---|---|---|---|---|---|---|---|
| VGG-13_w_BN | 46.81 | 78.05 | 70.97 | 91.87 | 69.77 | 73.98 | 66.03 | 62.72 | 68.11 | 77.86 | 57.46 | 133.1M |
| VGG-16_w_BN | 44.63 | 76.34 | 68.36 | 91.33 | 64.86 | 72.23 | 63.92 | 57.03 | 66.90 | 72.16 | 58.46 | 138.4M |
| VGG-19_w_BN | 43.00 | 75.19 | 66.33 | 91.02 | 63.24 | 71.29 | 56.17 | 56.09 | 62.55 | 70.43 | 59.88 | 143.7M |
| ResNet-18 | 44.34 | 77.17 | 69.56 | 91.78 | 66.03 | 74.35 | 61.99 | 65.18 | 65.23 | 74.50 | 57.46 | 11.7M |
| ResNet-34 | 45.63 | 76.94 | 69.82 | 91.79 | 68.15 | 72.10 | 61.75 | 60.82 | 70.12 | 74.22 | 59.64 | 21.8M |
| ResNet-50 | 44.15 | 77.27 | 69.20 | 92.24 | 68.17 | 73.37 | 63.34 | 60.94 | 66.72 | 68.03 | 60.79 | 25.6M |
| ResNet-101 | 44.86 | 77.18 | 69.77 | 91.30 | 69.05 | 73.32 | 64.63 | 63.10 | 67.74 | 68.24 | 60.76 | 44.5M |
| ResNet-152 | 45.74 | 77.77 | 70.09 | 92.54 | 68.63 | 73.11 | 64.37 | 57.03 | 67.05 | 79.71 | 58.30 | 60.2M |
| DenseNet-121 | 46.73 | 79.69 | 73.22 | 92.69 | 74.64 | 76.03 | 67.60 | 64.60 | 69.62 | 74.04 | 66.53 | 8.0M |
| DenseNet-161 | 48.72 | 80.01 | 74.57 | 92.09 | 73.78 | 76.72 | 66.04 | 68.92 | 72.56 | 79.19 | 67.26 | 28.7M |
| DenseNet-201 | 49.21 | 80.01 | 74.16 | 92.43 | 73.51 | 76.37 | 68.10 | 66.39 | 72.18 | 76.49 | 67.84 | 20.0M |
| Inception v3 | 44.74 | 77.19 | 69.70 | 91.78 | 70.83 | 71.72 | 62.82 | 61.98 | 66.51 | 70.52 | 61.47 | 27.2M |
| EfficientNet-B4 | 53.24 | 81.76 | 76.58 | 93.72 | 74.40 | 78.87 | 68.44 | 71.17 | 74.47 | <u>83.41</u> | 68.16 | 19.3M |
| EfficientNet-B5 | 53.67 | 82.22 | 76.78 | 94.08 | 74.41 | 79.66 | 68.80 | 71.69 | 74.73 | 83.17 | 67.70 | 30.4M |
| EfficientNet-B6 | 54.09 | 82.25 | 76.85 | 93.95 | 76.21 | 78.22 | 70.70 | 70.46 | 76.05 | 80.58 | 68.65 | 43.0M |
| EfficientNet-B7 | 54.92 | 82.99 | 77.54 | 94.10 | 76.82 | 80.85 | <u>71.38</u> | <u>72.58</u> | 75.05 | 80.85 | 68.65 | 66.3M |
| ViT-Base-16 | 37.90 | 71.75 | 61.40 | 88.86 | 61.10 | 68.97 | 48.74 | 49.42 | 54.94 | 74.04 | 45.17 | 86.6M |
| Swin-Transformer-Tiny | 53.38 | 81.24 | 75.17 | 93.15 | 75.62 | 75.39 | 69.53 | 67.64 | 76.67 | 73.71 | <u>69.67</u> | 28.3M |
| Swin-Transformer-Small | <u>58.21</u> | <u>83.89</u> | <u>78.30</u> | 94.40 | **78.26** | <u>81.44</u> | 67.96 | **75.38** | <u>76.78</u> | 82.56 | 69.65 | 49.6M |
| Swin-Transformer-Base | **58.49** | **84.00** | **78.87** | **94.49** | <u>77.89</u> | **81.59** | **72.12** | 70.52 | **78.85** | **84.56** | **70.92** | 87.8M |
| Swin-S + Swin-B + EfficientNet-B6 | 63.04 | 86.09 | 81.12 | **95.40** | 81.23 | 83.87 | 73.12 | 76.55 | 80.00 | 84.54 | 74.28 | 180.4M |
| Swin-S + Swin-B + EfficientNet-B7 | **63.33** | **86.39** | **81.74** | 95.25 | 81.13 | **84.41** | 73.63 | 77.53 | 80.91 | 85.06 | 76.03 | 203.7M |

TABLE 2  Summary of False Negatives (FNs), False Positives (FPs), and FNs/FPs for each class in this multi-label coral condition classification task.

|  | FNs | FPs | FNs/FPS |
|---|---|---|---|
| Healthy coral | 185 | 242 | 0.76 |
| Compromised coral | 374 | 250 | 1.50 |
| Dead coral | 388 | 363 | 1.07 |
| Rubble | 230 | 121 | 1.90 |
| Competition | 266 | 73 | 3.64 |
| Disease | 146 | 105 | 1.39 |
| Predation | 40 | 19 | 2.11 |
| Physical issues | 208 | 130 | 1.60 |

**Figure captions**

**Fig. 1** a) Overview of the study region and diving sites covered in the field survey, where HWB = Hin Wong Bay, AOM = Aow Mao, TTB = Tanote Bay, ALK = Aow Leuk, HNM = Hin Ngam, SKI = Shark Island, TCB = Taa Chaa Bay, CBK = Chalok Baan Kao Bay, SWP = Southwest Pinnacle. b) Example of underwater image containing branching coral colonies. c) Example of the close image taken within 40cm, showing the same branching coral colony with a cluster of *Drupella* snails as predators (marked by the arrow) on it.

**Fig. 2** Examples of representative patches from the dataset, where each column shows two examples per class. a) HLC = Healthy coral, b) CPC = Compromised coral, c) DDC = Dead coral, d) RBL = Rubble, e) CPT = Competition, f) DSE = Disease, g) PRD = Predation, and h) PHY = Physical issues. Since it is a multi-label classification task, some examples contain more than one class.

**Fig. 3** Framework illustrating the process of developing the deep learning model for the multi-label classification task. a) Training stage: Ground truth images were imported to the deep learning model, followed by the independent application of the binary cross-entropy loss function to each output channel. b) Inference stage: test images were input to three models simultaneously, and probabilities were averaged to determine the final prediction.

**Fig. 4** Examples show subtle, hard-to-be-identified features that indicate the presence of (a) competition and (b) physical issues; similarities between (c) rubble and (d) dead coral; inner variations within the competition class, where (e) images of four different types of macro-algae with various morphological traits and (f) images of different sponges (left two) and cyanobacteria (right two).